# Scalable Attentive Sentence-Pair Modeling via Distilled Sentence Embedding

**Oren Barkan**[*1], **Noam Razin**[*12], **Itzik Malkiel**[12], **Ori Katz**[13], **Avi Caciularu**[14], **Noam Koenigstein**[12]

[1]Microsoft, [2]Tel Aviv University, [3]Technion, [4]Bar-Ilan University

### Abstract

Recent state-of-the-art natural language understanding models, such as BERT and XLNet, score a pair of sentences ($A$ and $B$) using multiple *cross-attention* operations – a process in which each word in sentence $A$ attends to all words in sentence $B$ and vice versa. As a result, computing the similarity between a query sentence and a set of candidate sentences, requires the propagation of all query-candidate sentence-pairs throughout a stack of cross-attention layers. This exhaustive process becomes computationally prohibitive when the number of candidate sentences is large. In contrast, sentence embedding techniques learn a sentence-to-vector mapping and compute the similarity between the sentence vectors via simple elementary operations. In this paper, we introduce Distilled Sentence Embedding (DSE) – a model that is based on knowledge distillation from cross-attentive models, focusing on sentence-pair tasks. The outline of DSE is as follows: Given a cross-attentive *teacher* model (e.g. a fine-tuned BERT), we train a sentence embedding based *student* model to reconstruct the sentence-pair scores obtained by the teacher model. We empirically demonstrate the effectiveness of DSE on five GLUE sentence-pair tasks. DSE significantly outperforms several ELMO variants and other sentence embedding methods, while accelerating computation of the query-candidate sentence-pairs similarities by several orders of magnitude, with an average relative degradation of 4.6% compared to BERT. Furthermore, we show that DSE produces sentence embeddings that reach state-of-the-art performance on universal sentence representation benchmarks. Our code is made publicly available at https://github.com/microsoft/Distilled-Sentence-Embedding.

## 1. Introduction

The emergence of self-attentive models such as the Transformer (Vaswani et al. 2017), GPT (Radford et al. 2018), BERT (Devlin et al. 2019) and XLNet (Yang et al. 2019) significantly advanced the state-of-the-art in various linguistic tasks such as machine translation (Vaswani et al. 2017), sentiment analysis (Socher et al. 2013), question answering (Rajpurkar et al. 2016) and sentence similarity (Dolan and Brockett 2005; Cer et al. 2017). These models are built upon a stack of self-attention layers that enable each word to attend other words in a sentence.

In the latest models such as BERT and XLNet, self-attention is applied in a bidirectional manner. This is different from conventional language models (Collobert et al. 2011), in which each word in a sentence is conditioned solely on its preceding words. In addition, the architectures in (Devlin et al. 2019; Yang et al. 2019) support sentence-pair input, endowing these models with the ability to infer sentence similarity. However, this capability entails a non-negligible computational cost. In these models, scoring sentence-pairs involves a *cross-attention* (CA) operation in which each word in a sentence $A$ attends to all words in a sentence $B$ and vice versa (excluding the fact that each word attends to all other words in the same sentence as well). Moreover, CA is repeatedly applied in a cascade throughout a stack of multi-head attention layers (Vaswani et al. 2017). This CA entanglement is a double-edged sword. On the one hand, it is allegedly a key property that pushes forward the state-of-the-art, computing similarity between sentences by analyzing the relations between individual words $a \in A$ and $b \in B$. On the other hand, it entails an excessively demanding inference phase in terms of time and computational power.

The computational bottleneck that is imposed by CA severely affects the inference phase in ranking and retrieval tasks. Assume a CA model $T$ and a set of candidates $X$ that contains $N$ sentences. The task is to retrieve the topmost similar sentences in $X$ w.r.t. a new query sentence $q$. A naïve solution is to compute the similarity between each sentence $x \in X$ and $q$, which amounts to $N$ applications of $T$ (scoring each sentence-pair $(q, x)$ using $T$). In other words, the propagation of the entire candidates set $X$ through $T$ is necessary to produce the similarity scores w.r.t. a single query $q$.

A second problem with CA models is the fact that they are not trained to produce sentence embeddings w.r.t. the task at hand. While several types of heuristics can be employed to produce sentence embedding (e.g., summing the several last hidden token representations, using the CLS hidden token (Devlin et al. 2019) as a sentence representation, etc.), none of them are truly justified. These operations are employed after the training phase is over and are not directly related to the original training objective. This problem is a key differentiator between CA models and other models (Conneau et al. 2017; Subramanian et al. 2018) that inherently support sentence embedding.

---

[*]Equal contribution

In this paper, we present Distilled Sentence Embedding (DSE), a model for learning a sentence embedding via Knowledge Distillation (Hinton et al. 2014) from CA models. The essence of DSE is as follows: Given a trained CA *teacher* model and a *student* model. We train the student model to map sentences to vectors in a latent space, in which the application of a low-cost similarity function approximates the similarity score obtained by the teacher model for the corresponding sentence-pair. Specifically, DSE employs a pairwise training procedure in which each pair of sentences $(A, B)$ and score $T_{AB}$ (that is obtained by the teacher model for the specific sentence-pair) is treated as a training example. The student model consists of parametric embedding and similarity functions. The embedding function maps the sentences $A$ and $B$ to vectors, on which the similarity function is applied to produce a similarity score $S_{AB}$. Finally, using a loss function, we compare between $S_{AB}$ and $T_{AB}$.

During the training phase, the student model parameters (that includes both the embedding and similarity functions) are learned via stochastic gradient descent w.r.t. a loss function that compares the student output score $S_{AB}$ to the teacher model score $T_{AB}$. In the inference phase, the student model maps an input sentence-pair to a vector-pair using the embedding function and then computes the vector-pair similarity score using the similarity function. DSE essentially performs a disentanglement that enables the precomputation of the candidate sentence embeddings in advance. As a result, for ranking and retrieval tasks, the computational complexity of a query reduces to a single application of the student model to the query sentence $q$, followed by $N$ applications of the low-cost similarity function (for each vector-pair).

We evaluate DSE on five sentence-pair tasks from the GLUE benchmark (Wang et al. 2018). Empirical results show that DSE significantly outperforms other sentence embedding methods as well as several attentive ELMO (Peters et al. 2018) variants, while providing average relative degradations of 4.6% and 3.1% compared to BERT-Large and BERT-Base, respectively. We further analyze the quality of sentence embeddings produced by DSE on standard universal sentence representation benchmarks (Conneau and Kiela 2018). In this setting, DSE is initially pre-trained on a surrogate task. Then, general purpose sentence representations are extracted from the model and evaluated on downstream tasks. The obtained embeddings are competitive with current top performing approaches.

Our main contributions are as follows: 1) We present DSE, a novel sentence embedding model that is supervised by the original sentence-pair similarity scores of state-of-the-art CA models. 2) We show that DSE, as a general purpose sentence embedding method, reaches state-of-the-art performance on standard universal sentence representation benchmarks. 3) DSE significantly speeds-up the computation of online and offline query-candidate similarities, posing a practical solution for mass production systems at the cost of a relatively small degradation in performance.

## 2. Related Work

There have recently appeared an increasing number of studies suggesting usage of general language representation models for natural language understanding tasks. Among the most promising techniques, the unsupervised fine-tuning approach has been shown to be effective on many sentence-level tasks (Dai and Le 2015; Howard and Ruder 2018; Radford et al. 2018). This technique uses a sentence encoder to produce contextual token representations. The encoder training procedure is composed of two phases: (1) unsupervised training on unlabeled text, and (2) fine-tuning for supervised downstream tasks. The unsupervised training allows the model to learn most of the parameters in advance, leaving only few parameters to be learned from scratch during fine-tuning.

More recently, BERT (Devlin et al. 2019) has emerged as a powerful method that has achieved state-of-the-art results in various sentence or sentence-pair language understanding tasks from the GLUE benchmark (Wang et al. 2018), including sentiment analysis (Socher et al. 2013), paraphrase identification (Williams et al. 2017) and semantic text similarity (Cer et al. 2017). Liu et al. (Liu et al. 2019), introduce Multi-Task Deep Neural Network (MT-DNN), which extends BERT by learning text representations across multiple natural language understanding tasks. In sentence-pair tasks, both BERT and MT-DNN require feeding both sentences together as a single input sequence. While other techniques, such as (Conneau et al. 2017; Subramanian et al. 2018), suggest extracting a feature vector for each sentence separately via an embedding function, followed by a relatively low cost similarity function which produces a similarity score for the vector-pair.

The problem of reducing the computational burden of neural networks at inference time has attracted considerable attention in the literature. Hinton et al. (Hinton et al. 2014), introduced Knowledge Distillation (KD) as a framework for model compression, where knowledge from a large model is used for training a simple model, by following a teacher-student paradigm. Specifically, the method leverages the probabilities produced by a teacher model for training a simple student model, by teaching the student to predict both the true labels and the output probabilities of the teacher.

In the context of natural language understanding, Liu et al. (Liu et al. 2019) propose to distill knowledge from an ensemble of MT-DNN models (teachers), into a single MT-DNN model (a student). The authors show that by leveraging KD, a student MT-DNN model significantly outperforms the original MT-DNN (Liu et al. 2019) on various linguistic tasks.

Different from other KD studies, our method focuses on distilling knowledge from a CA model (the teacher), into a sentence embedding model that solely relies on self-attention (the student). Specifically, we leverage KD for training a student BERT model to bypass BERT's requirement of feeding sentence-pairs as a unified sequence.

Excluding such an intrinsic operation from the student might hinder its ability to perfectly reconstruct its teacher's knowledge. However, such a property would ease the adoption of BERT in other tasks, such as ranking and information retrieval, which require exhaustive computations across many documents or paragraphs in a given dataset.

## 3. Distilled Sentence Embedding (DSE)

In this section, we present the problem setup and describe the DSE model in detail.

### 3.1 Problem Setup

Let $\mathcal{W} = \{w_i\}_{i=1}^{w}$ be the vocabulary of all supported tokens. We define $Y$ to be the set of all possible sentences that can be generated using the vocabulary $\mathcal{W}$.

Let $T: Y \times Y \to \mathbb{R}$ be the teacher model (e.g., a fine-tuned BERT model). $T$ receives a sentence-pair $(y, z) \in Y \times Y$ and outputs a similarity score $T_{yz} \triangleq T(y, z)$. Note that $T$ is not necessarily a symmetric function.

Let $\psi, \phi: Y \to \mathbb{R}^d$ be sentence embedding functions that embed a sentence $y \in Y$ in a $d$-dimensional latent vector space. The usage of different sentence embedding functions, $\psi$ and $\phi$, is due to the fact that $T$ is not necessarily a symmetric function. For example, in BERT, the sentences $A$ and $B$ are associated with different segment embeddings. Therefore, $\psi$ and $\phi$ play a similar role as the common context and target representations that appear in many neural embedding methods (Barkan 2017; Barkan and Koenigstein 2016; Mikolov et al. 2013; Mnih and Hinton 2009).

Let $f: \mathbb{R}^d \times \mathbb{R}^d \to \mathbb{R}$ be a (parametric) similarity function. $f$ scores the similarity between sentence embeddings that are produced by $\psi$ and $\phi$. Then, the student model $S: Y \times Y \to \mathbb{R}$ is defined as

$$S_{yz} \triangleq f(\psi(y), \phi(z)). \qquad (1)$$

Given a set of paired training sentences $X = \{(y_i, z_i)\}_{i=1}^{N}$, our goal is to train the student model $S$ such that for all $(y, z) \in X$, its similarity score $S_{yz}$ approximates the teacher model's score $T_{yz}$ with a high accuracy. To this end, we propose to learn the student model parameters via a pairwise training procedure, which is explained in Section 3.2.

Note that in some sentence-pair tasks the teacher model's codomain is multidimensional. For example, the MNLI (Williams et al. 2017) task is to predict whether the relation between two sentences is neutral, contradictory or entailment. In this case, the codomain of the teacher model $T$ is $\mathbb{R}^3$ and hence the codomain of the similarity function $f$ (and the student model $S$) is $\mathbb{R}^3$ as well.

### 3.2 Pairwise Training

In pairwise training, we define a loss function $\mathcal{L}: \mathbb{R} \times \mathbb{R} \to \mathbb{R}$ and train $S$ to minimize $\mathcal{L}(S_{yz}, T_{yz})$ in an end-to-end fashion. Specifically, given a sentence-pair $(y, z) \in X \times X$, we compute the embeddings $\psi(y)$ and $\phi(z)$ for the sentences $y$ and $z$, respectively. Then, the similarity score $S_{yz}$ is computed using the similarity function $f$ according to Eq. (1).

Note that $\mathcal{L}$ can be either a regression or classification loss depending on the task at hand. Moreover, $\mathcal{L}$ can be trivially extended to support multiple teacher models. In (Hinton et al. 2014) the authors suggest using two teacher models $T$ and $R$, where $R$ is simply the ground truth labels as follows

$$\mathcal{L}_{yz} = \alpha l_{dstl}(S_{yz}, T_{yz}) + (1 - \alpha) l_{lbl}(S_{yz}, R_{yz}) \qquad (2)$$

where $\alpha \in [0,1]$ is a hyperparameter that controls the relative amount of supervision that is induced by $T$ and $R$. In this case, the student model is simultaneously supervised by $T$ and $R$. Note that in general, the distillation loss $l_{dstl}$ and the ground truth label loss $l_{lbl}$ are not restricted to be the same loss function (as shown in Section 3.5). The DSE model is illustrated in Fig. 1.

### 3.3 The Teacher Model

The teacher model $T$ is implemented as a BERT-Large model from (Devlin et al. 2019), consisting of 24 encoder layers that each employ a self-attention mechanism. For a sentence-pair input, $T$ employs CA between the two sentences. The teacher model is initialized to the pre-trained version from (Devlin et al. 2019) and then fine-tuned according to each specific sentence-pair task.

After the fine-tuning phase, we compute the score $T_{yz}$ for a sentence-pair $(y, z)$ by propagating a unified representation of the sentence-pair throughout $T$, as done in (Devlin et al. 2019). The score is then extracted from the output layer, which is placed on top of the last hidden representation of the CLS token. Note that $T_{yz}$ is set to the logit value (before the softmax / sigmoid activation).

It is important to emphasize that DSE is not limited to BERT as a teacher model. For example, we could use the exact same method with an XLNet (Yang et al. 2019) teacher. The choice of BERT is mainly due to its prevalence in the natural language understanding community.

### 3.4 The Student Model

The teacher model is based on BERT, which is not symmetric due to its use of different segment embeddings for input sentences. Yet, to refrain from doubling the number of parameters, we decide to implement a symmetric (Siamese) student model by learning a single mutual embedding function $\psi$ ($= \phi$). The embedding function $\psi$ is implemented using a BERT-Large model that operates on a single sentence (using only the segment embedding A) and outputs a vector representation. Specifically, given a sentence $y$, we first add the CLS token to the beginning of $y$ and a SEP token to the end, before feeding $\psi$ with the resulted representation. Then, we compute the average pooling operation across the hidden tokens for each of the last four encoder layers' outputs. Experimentally, we observed degradations in performance when including the CLS token in the pooling of hidden tokens, thus, it is excluded from it. We attribute this to the fact that during pre-training the CLS token representation is used for encoding information across two sentences for the Next Sentence Prediction task (Devlin et al. 2019), and is therefore not well suited for representing a single sentence. The pooling operation produces four 1024-dimensional vectors (one for each encoder layer) that are then concatenated to form a 4096-dimensional representation as the final sentence embedding (hence $d = 4096$).

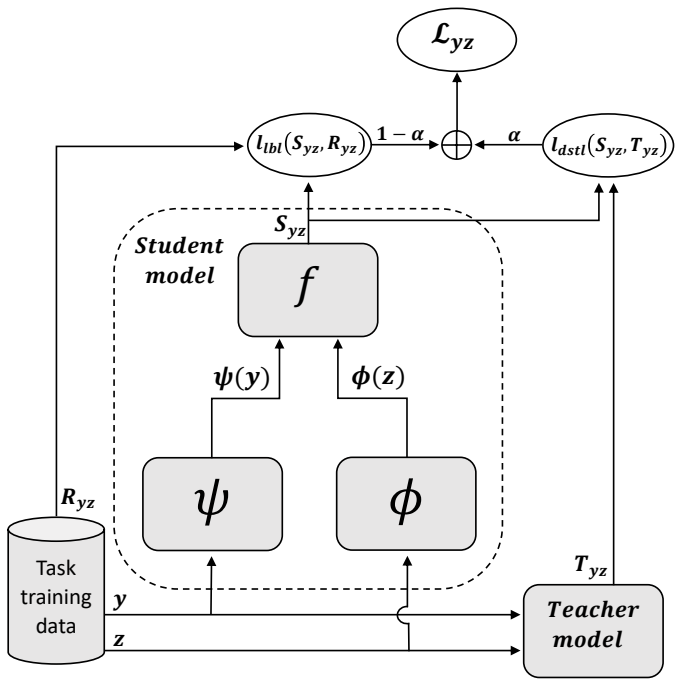

Figure 1. A schematic illustration of the DSE model.

Inspired by (Wang et al. 2018), we use the similarity function

$$f(u,v) = w^T ReLU(Wh) \quad (3)$$

where $h = [u, v, u \circ v, |u - v|] \in \mathbb{R}^{16384}$ ($\circ$ stands for the Hadamard product), $W \in \mathbb{R}^{512 \times 16384}$ and $w \in \mathbb{R}^{512}$. Both $W$ and $w$ are learnable parameters. Note that $u, v \in \mathbb{R}^{4096}$ are the sentences' representations that are produced by the embedding function $\psi$.

Like the teacher model, $\psi$ is initialized to the pre-trained version of BERT-Large from (Devlin et al. 2019). Note that we could initialize $\psi$ to the fine-tuned teacher model, however in our initial experiments we found it to perform worse.

### 3.5 The Loss Function

We implement a loss function according to Eq. (2) - a linear combination of the distillation and label losses. The distillation loss term is set to the L2 loss

$$l_{dstl}(S_{yz}, T_{yz}) = l_{L2}(S_{yz}, T_{yz}) = \|S_{yz} - T_{yz}\|_2^2.$$

The motivation behind this choice is the analysis from (Hinton et al. 2014), where it is shown that for high temperature values, minimization of the cross-entropy loss over the softmax outputs, is equivalent to minimizing L2 loss over the logits (before applying softmax). Indeed, our initial experiments revealed that using the L2 loss on the logits produces superior distillation results.

The label loss is set according to the task at hand: For a multiclass classification task we set

$$l_{lbl}(S_{yz}, R_{yz}) = l_{cce}(\rho(S_{yz}), R_{yz})$$

where $R_{yz} \in \{0,1\}^n$ is a one-hot vector, $\rho(S_{yz}) \in [0,1]^n$ is a discrete probability distribution obtained by applying the softmax function $\rho$, and $l_{cce}(a, b) = -\sum_{i=1}^{n} b_i \log a_i$ is the categorical cross entropy loss. For a binary classification task, we use the same loss with $n = 2$. For a regression task we set $l_{lbl}(S_{yz}, R_{yz}) = l_{L2}(S_{yz}, R_{yz})$, where $R_{yz} \in \mathbb{R}$.

## 4. Experimental Setup and Results

We evaluate DSE in two different settings: First, task specific embeddings for sentence-pair tasks, where the whole model is trained in an end-to-end fashion and evaluated on a specific dataset. Second, universal sentence representations generation, in which the model is pre-trained to produce general purpose sentence embeddings. In addition, we report empirical results that showcase the efficiency of DSE in computing sentence-pair similarities compared to BERT.

### 4.1 Sentence-Pair Modeling

For sentence-pair tasks, our evaluation includes several datasets from the GLUE benchmark: MRPC (Dolan and Brockett, 2005), MNLI (Williams et al., 2018), QQP, QNLI (Wang et al., 2018), and STS-B (Cer et al., 2017). These datasets represent different tasks that revolve around modeling and scoring sentence-pairs. MRPC, STS-B, and QQP focus on semantic similarity of phrases or questions, MNLI is a natural language inference (NLI) benchmark, and lastly, QNLI is a question answering dataset. We refer to (Wang et al. 2018) for a detailed description of these datasets.

#### 4.1.1 Models and Hyperparameter Configuration

The models that participate in the evaluation are as follows:

| Model | MNLI | QQP | QNLI | MRPC | STS-B | AVG | Degradation compared to BERT-Large | Improvement obtained by DSE ($\alpha = 0.5$) |
|---|---|---|---|---|---|---|---|---|
| Cross Attentive Models | | | | | | | | |
| BERT-Large | 86.7/85.9 | 72.1/89.3 | 92.7 | 89.3/85.4 | 87.6/86.5 | 86.82 | 0% | -4.6% |
| BERT-Base | 84.6/83.4 | 71.2/89.2 | 90.2 | 88.9/84.8 | 87.1/85.8 | 85.54 | 0.9% | -3.1% |
| ELMO + Attn (MT) | 74.1/74.5 | 63.1/84.3 | 79.8 | 84.4/78.0 | 74.2/72.3 | 76.45 | 11.9% | 8.3% |
| ELMO + Attn (ST) | 76.9/76.7 | 66.1/86.5 | 76.7 | 80.2/68.8 | 55.5/52.5 | 71.66 | 17.4% | 15.6% |
| Sentence Embedding Models | | | | | | | | |
| GenSen | 71.4/71.3 | 59.8/82.9 | 78.6 | 83.0/76.6 | 79.3/79.2 | 76.07 | 12.3% | 8.9% |
| DSE ($\alpha = 1$) | 80.3/79.4 | 68.4/86.8 | 86.1 | 86.8/80.8 | 86.8/86.1 | 82.76 | 4.7% | 0.1% |
| DSE ($\alpha = 0.5$) | 80.9/80.4 | 68.5/86.9 | 86.0 | 86.7/80.7 | 86.4/85.8 | 82.83 | 4.6% | 0% |
| DSE ($\alpha = 0$) | 79.7/79.0 | 67.0/86.8 | 84.8 | 86.5/79.9 | 87.0/86.5 | 82.2 | 5.3% | 0.8% |
| DSE (Frozen $\psi$, $\alpha = 1$) | 69.3/69.9 | 62.5/81.4 | 76.9 | 86.5/79.9 | 73.3/73.0 | 74.96 | 13.6% | 10.5% |

Table 1: A comparison between DSE to other models across different test sets. For MNLI, accuracy is reported for matched / mismatched test sets. For QQP and MRPC, F1/accuracy scores are reported. For QNLI, accuracy is reported. For STS-B, Pearson / Spearman correlation coefficients are reported. AVG column presents the average score across all datasets, where each dataset's score is the mean of its one or two reported scores. Degradation and improvement columns present the relative degradation compared to BERT-Large and the relative improvement obtained by DSE ($\alpha = 0.5$) over each model (reported in percentages), respectively.

**BERT-Large:** This is the BERT-Large model from (Devlin et al. 2019). This model is also used as a teacher model. Results are reported from (Devlin et al. 2019).

**BERT-Base:** This is the BERT-Base model from (Devlin et al. 2019). Results are reported from (Devlin et al. 2019).

**DSE:** This is our proposed model from Section 3. We consider three variants of DSE that differ by the parameter values of $\alpha \in \{0, 0.5, 1\}$ which controls the amount of distillation. For all datasets we set the distillation loss $l_{dstl} = l_{L2}$. For QQP, MRPC, QNLI and MNLI we set the label loss $l_{lbl} = l_{cce}$. Specifically, for MNLI we further used $w \in \mathbb{R}^{3 \times 512}$ in Eq. (3) to support a 3-dimensional output. For STS-B, we set $l_{lbl} = l_{L2}$. We used the Adam optimizer (Kingma and Ba 2014) with minibatch size of 32 and a learning rate of 2e-5, except for STS-B, where we used a learning rate of 1e-5. The models were trained for 8 epochs. The best model was selected based on the dev set.

**DSE (Frozen $\psi$):** We trained another version of DSE in which $\psi$ is frozen. Since $\psi$ is implemented as BERT (Section 3.4), we further want to investigate the actual benefit from fine-tuning $\psi$ w.r.t. the task at hand. Therefore, we present results for a DSE version in which $\psi$ is not fine-tuned. Note that the parametric similarity function is still learned in this version.

**ELMO + Attn:** This is the BiLSTM + ELMO, Attn model from (Wang et al. 2018). It comes in two variants: Single-Task (ST) and Multi-Task (MT) Training. The results are reported taken from (Wang et al. 2018).

**GenSen:** Since DSE is a sentence embedding model, we further compare its performance with GenSen (Subramanian et al. 2018), which is the best performing sentence embedding model from (Wang et al. 2018). The results are taken from (Wang et al. 2018).

### 4.1.2 Sentence-Pair Tasks Results

Table 1 presents the results for each combination of model and dataset. In addition, we provide the average score that is computed across the datasets for each model (AVG column). The last two columns present the relative degradation compared to BERT-Large and the relative improvement obtained by DSE ($\alpha = 0.5$) over each model (reported in percentages).

First, we compare between the four DSE variants. We see that for MNLI, QNLI, MRPC and QQP, enabling distillation ($\alpha \in \{0.5, 1\}$) slightly improves upon using $\alpha = 0$. However, on STS-B, distillation seems to hurt performance. We attribute the degradation to the fact that STS-B is a regression task and therefore the ground truth labels are already provided in a resolution that is finer than binary values. Lastly, we see that the frozen version of DSE performs much worse than all other DSE variants. This is evidence for the importance of fine-tuning $\psi$, which further confirms that a naïve use of pre-trained BERT for sentence embedding produces relatively poor results, in some cases. Therefore, we conclude that the distilled version of DSE ($\alpha \in \{0.5, 1\}$) performs the best. From now on, we focus on a comparison between the $\alpha = 0.5$ version of DSE and the other models.

Next, we turn to consider the performance gaps between DSE and BERT. Recall that DSE is supervised by BERT-Large and hence the performance gaps between the two models quantifies the ability of the former to reconstruct the latter's scores. We see that the largest and smallest relative degradations occur on the MNLI and STS-B datasets, respectively. Overall, DSE results in an average relative degradations of 4.6% and 3.1% compared to BERT-Large and BERT-Base, respectively. We attribute these degradations to the fact that DSE lacks the CA mechanism that exists in BERT, which seems to capture important information that yields further improvements.

| Model | MR | CR | SUBJ | MPQA | SST | TREC | MRPC | SICK-R | SICK-E | STS-B | AVG |
|---|---|---|---|---|---|---|---|---|---|---|---|
| GenSen | 82.5 | 87.7 | 94.0 | 90.9 | 83.2 | 93.0 | 84.4/78.6 | 0.888 | 87.8 | 78.9/78.6 | 86.8 |
| InferSent | 81.1 | 86.3 | 92.4 | 90.2 | 84.6 | 88.2 | 83.1/76.2 | 0.884 | 86.3 | 75.8/75.5 | 85.2 |
| BERT-Large | 83.5 | 88.8 | 95.5 | 89.1 | 87.1 | 93.2 | 83.5/76.4 | 0.838 | 82.2 | 68.4/68.3 | 85.1 |
| DSE ($\alpha = 0.5$) | 83.6 | 90.2 | 93.6 | 89.8 | 91.0 | 91.8 | 83.8/77.9 | 0.856 | 86.7 | 70.7/71.4 | 86.4 |
| DSE ($\alpha = 0$) | 83.1 | 89.8 | 93.1 | 89.4 | 88.3 | 92.0 | 81.8/76.2 | 0.847 | 86.1 | 73.1/74.1 | 85.9 |

Table 2: Universal sentence embedding benchmarks results. The evaluation results are of linear models trained over each of the model's sentence representations. The results for Gensen and InferSent are taken from their respective papers. We report the F1/accuracy scores for MRPC, Pearson correlation for SICK-R, Pearson/Spearman correlations for STSB, and accuracy for the rest. AVG column presents the average score across all datasets, where each dataset's score is the mean of its one or two reported scores.

Next, we turn to compare the performance of DSE against the ELMO + Attn variants, which are the best performing models from (Wang et al. 2018). First, we see that DSE significantly outperforms ELMO + Attn (ST) across all datasets with an average relative improvement of 15.6%. This is despite the fact ELMO + Attn employs CA operations. However, it is important to note that the CA mechanism in ELMO + Attn (Appendix B.1 in (Wang et al. 2018)) is substantially different from the one that exists in BERT (Devlin et al. 2019), which is based on self-attention (Vaswani et al. 2017). Moreover, DSE provides an average relative improvement of 8.3% compared to ELMO + Attn (MT), even though DSE is trained in a single-task manner.

Finally, we turn to compare between DSE and GenSen (Subramanian et al. 2018), which is reported to perform the best among all other sentence embedding models in (Wang et al. 2018). Table 1 shows that DSE outperforms GenSen across all datasets, providing an average relative improvement of 8.9%. Yet, it is important to note that although GenSen is pre-trained on MNLI and SNLI (Bowman et al. 2015) datasets, it is not fine-tuned w.r.t. the rest of the tasks from Section 4.1. Similar analysis holds for other currently existing pre-trained sentence embedding methods, such as (Kiros et al. 2015; Nie, Bennett, and Goodman 2017; Conneau et al. 2017), with DSE obtaining an even higher relative improvement. Therefore, we omit their results from Table 1.

To the best of our knowledge, these results place DSE as the best performing task specific fine-tuned sentence embedding model on GLUE sentence-pair tasks.

## 4.2 Universal Sentence Embeddings

We further evaluate DSE by examining its applicability for producing universal sentence embeddings. In this setup, the model is initially pre-trained on one or more surrogate tasks. Then, the learned model is used to generate sentence embeddings, that are evaluated on various downstream tasks in a separate procedure that does not further update the pre-trained model.

Choosing a suitable pre-training task and dataset is crucial for learning representations that are meaningful for multiple tasks. Current top performing approaches mainly vary in this aspect, suggesting both supervised and unsupervised techniques. Following (Conneau et al. 2017), we opt for pre-training DSE on the AllNLI (MNLI + SNLI) dataset. Sentence embeddings are then extracted from the student model and evaluated on standard benchmarks using the SentEval toolkit (Conneau and Kiela 2018). Both pre-training and sentence embedding generation are done as described in Section 3. We refer to (Conneau and Kiela 2018) for a detailed description of the datasets and the evaluation protocol.

### 4.2.1 Downstream Tasks Results

For each sentence embedding method and dataset included in the evaluation, Table 2 contains the results of a shallow linear model trained on top of the precomputed embeddings. We report results for our approach with $\alpha = 0.5$, which showed the most promising performance in Section 4.1.2, and compare it to the current state-of-the-art methods: Infersent (Conneau et al. 2017) and Gensen. Additionally, we include a comparison to a DSE variant without distillation ($\alpha = 0$), and to sentence embeddings that are extracted from a pre-trained BERT-Large model using the procedure described in Section 3.4.

As can be seen in Table 2, BERT-Large embeddings reach competitive results on several datasets to both InferSent and GenSen. Significant improvements are observed mostly for sentiment analysis related datasets. In contrast, on STS-B (semantic similarity), SICK-R, and SICK-E (NLI), BERT-Large embeddings are subpar compared to InferSent and GenSen, which are pre-trained directly on NLI datasets. Furthermore, recall that BERT is not explicitly trained to generate sentence embeddings, possibly explaining the downfalls in some of the tasks.

We now turn to compare DSE with the other baselines. As in the sentence-pair tasks evaluation, using DSE with $\alpha = 0.5$ improves upon the non-distilled variant ($\alpha = 0$), outperforming it on 8 of the 10 benchmarks. Specifically, substantial gains are obtained on SST and MRPC, demonstrating the effectiveness of knowledge distillation. Therefore, from now on, DSE relates to the $\alpha = 0.5$ model.

DSE significantly outperforms BERT-Large embeddings on sentiment analysis and NLI tasks, obtaining a relative improvement of 4.4% and 5.4% on SST and SICK-E respectively. The improved results on SICK-E, as well as on SICK-R, are straightforwardly explained by the pre-training procedure of DSE, which is done on NLI datasets. An interesting byproduct is the improvement on sentiment analysis benchmarks, suggesting there is a strong connection between the tasks. In total, DSE outperforms BERT-Large embeddings on 8 datasets, emphasizing the importance of further fine-tuning BERT in a sentence embedding oriented manner.

| Model | Speedup factor | Time | Max batch size | Time / Max batch size |
|---|---|---|---|---|
| BERT-Large | 1 | 9.66hr | 300 | 10.43s |
| DSE ($\psi$ phase) | - | 34.73s | 300 | 10.43s |
| DSE ($f$ phase) | - | 2.48s | 200k | 0.49s |
| DSE | **934** | 37.21s | - | - |

Table 3: Time comparison between DSE and BERT-Large for an offline computation of 1M sentence-pairs similarities for a catalog of 1000 sentences.

| Model | Speedup factor | Time |
|---|---|---|
| BERT-Large | 1 | 58m |
| DSE ($\psi$ phase) | - | 0.052s |
| DSE ($f$ phase) | - | 0.204s |
| DSE | **13594** | 0.256s |

Table 4: Time comparison between DSE and BERT-Large for an online computation of sentence similarities between a query sentence and a catalog of 100K sentences.

Finally, we turn to compare DSE to current state-of-the-art methods: InferSent and GenSen. DSE is competitive with both, outperforming them on 7 and 3 datasets, respectively. We attribute these impressive results, especially on sentiment analysis benchmarks, to the larger transformer architecture and robust BERT pre-training that the DSE student model is based on. On these datasets, BERT-Large improves upon InferSent and GenSen as well, though by a smaller margin, further suggesting this notion. On average, GenSen remains the current top performing method, slightly outperforming DSE. GenSen utilizes an extensive pre-training phase that includes multiple supervised and unsupervised tasks. The fact that DSE surpasses GenSen by a large margin on some datasets, yet underperforms on others, implies that multitask pre-training results in good generalization on various tasks, while the specialized NLI pre-training can suffer on tasks that it is less correlated with. Overall, DSE is shown to provide state-of-the-art results.

### 4.3 Computational Efficiency Evaluation

In this section, we report computation times that were measured for DSE and BERT-Large. We conducted two experiments on a single NVIDIA V100 32GB GPU using PyTorch. The first experiment is designed to simulate an offline computation of a pairwise sentence similarity matrix. To this end, we compute the 1M optional sentence-pair similarities between 1000 sentences. For BERT-Large, we simply performed 1M forward passes with a maximal batch size of 300. This operation took ~9.6 hours. For DSE, we first computed the 1000 sentence embeddings using $\psi$, which amounts to 1000 forward passes of BERT-Large with the same batch size of 300. This operation took ~35 seconds. Then, we computed the 1M pairwise similarities between the sentence embeddings using $f$ with a maximal batch size of 200K. This operation took ~2 seconds.

Table 3 summarizes the results. We see that DSE provides a computation time that is 934 times faster than BERT-Large: 37 seconds vs. 9.6 hours. Therefore, we conclude that for large datasets that contain tens of thousands of sentences, computing the sentence-pair similarity matrix using BERT-Large becomes infeasible, while DSE remains a practical solution. In addition, we see that $f$ allows much larger batch sizes, compared to BERT-Large (200k vs. 300) with an average computation time per batch that is ~21 times faster.

The second experiment is designed to simulate a scenario of online query-candidate similarities computation. In this scenario, the task is to compute the similarities between a new query sentence to all the sentences in an existing catalog. It is assumed that the sentence embedding for all the sentences in the catalog are precomputed using $\psi$.

For BERT-Large, we ran 100K forward passes with a batch size of 300. This operation took ~58 minutes. For DSE, we compute the embedding for the query sentence using $\psi$ in 52 milliseconds and then compute the 100K query-candidate sentence similarities using $f$ in 204 milliseconds (all the similarities are computed in a single batch of size 100K as the maximal batch size is 200K, as reported in Table 2). The results are summarized in Table 4. We see that DSE provides a computation time that is 13.5K times faster than BERT-Large: 256 milliseconds seconds vs. 58 minutes.

The results in Tables 2 and 3 demonstrates the effectiveness of DSE: Obviously, using BERT-Large is impractical in both cases (offline and online), while DSE provides a practical alternative in the expense of a small relative degradation (4.6%) in quality compared to BERT-Large.

### 5. Conclusion

Computing sentence similarities via CA models such as BERT is impractical for large scale catalogs. To this end, we introduce DSE: a sentence embedding method that is based on knowledge distillation from CA models. DSE bypasses the need for CA operations, enabling precomputation of sentence representations for the existing catalog in advance, and fast query operations using a low-cost similarity function. We demonstrate the effectiveness of DSE on five sentence-pair tasks, where it is shown to outperform other sentence embedding methods as well as several attentive versions of ELMO. Furthermore, sentence embeddings produced by DSE provide state-of-the-art results on various benchmarks. Thus, DSE provides a practical solution for mass production systems, allowing sentence similarities computation times that are several orders of magnitude faster compared to BERT-Large, at the cost of a small relative degradation.

# References

Barkan, O. 2017. Bayesian neural word embedding. In *AAAI*, 3135–3143.

Barkan, O., and Koenigstein, N. 2016. Item2vec: Neural item embedding for collaborative filtering. In *IEEE MLSP, 2016*.

Bowman, S. R.; Angeli, G.; Potts, C.; and Manning, C. D. 2015. A large annotated corpus for learning natural language inference. *EMNLP* 632–642.

Cer D.; Diab M.; Agirre E.; Lopez-Gazpio I. and Lucia S. 2017 SemEval-2017 task 1: Semantic textual similarity multilingual and crosslingual focused evaluation. In *Proceedings of SemEval-2017*, pages 1–14, Vancouver, Canada.

Collobert, R.; Weston, J.; Bottou, L.; Karlen, M.; Kavukcuoglu, K.; and Kuksa, P. 2011. Natural language processing (almost) from scratch. *Journal of Machine Learning Research 12 (Aug)*:2493–2537.

Conneau, A.; Kiela, D.; Schwenk, H.; Barrault, L.; and Bordes, A. 2017. Supervised learning of universal sentence representations from natural language inference data. In *EMNLP*, 670–680.

Conneau, A., and Kiela, D. 2018. Senteval: An evaluation toolkit for universal sentence representations. *LREC*.

Dai, A. M., and Le, Q. V. 2015. Semi-supervised sequence learning. In *Advances in Neural Information Processing Systems*, 3061–3069.

Devlin J.; Chang M-W; Lee K; and Toutanova K. 2019. BERT: Pre-training of deep bidirectional transformers for language understanding. In *NAACL*.

Dolan, W. B. and Brockett, C. 2005. Automatically constructing a corpus of sentential paraphrases. In *IWP@IJCNLP*.

Hinton, G.; Vinyals, O.; and Dean, J. 2014. Distilling the knowledge in a neural network. In *NIPS Workshop on Deep Learning and Representation Learning*. Hoffman, J.; Tzeng, E.; Park, T.; Zhu, J.; I

Howard, J., and Ruder, S. 2018. Universal language model fine-tuning for text classification. *arXiv preprint arXiv:1801.06146*.

Kingma, D. P., and Ba, J. 2014. Adam: A method for stochastic optimization. In *Proceedings of the 3rd International Conference on Learning Representations (ICLR)*.

Kiros, R.; Zhu, Y.; Salakhutdinov, R. R.; Zemel, R.; Urtasun, R.; Torralba, A.; and Fidler, S. 2015. Skip-thought vectors. In *Advances in Neural Information Processing Systems*, (pp. 3294-3302).

Liu, X.; He, P.; Chen, W.; and Gao, J. 2019. Multi-task deep neural networks for natural language understanding. *arXiv preprint arXiv:1901.11504*.

Liu, X.; He, P.; Chen, W.; and Gao, J. 2019. Improving Multi-Task Deep Neural Networks via Knowledge Distillation for Natural Language Understanding. *arXiv preprint arXiv:1904.09482*.

Mikolov, T.; Sutskever, I.; Chen, K.; Corrado, G. S.; and Dean, J. 2013. Distributed representations of words and phrases and their compositionality. In *NIPS*.

Mnih, A., and Hinton, G. E. 2009. A scalable hierarchical distributed language model. In *NIPS*.

Nie, A.; Bennet, E. D.; and Goodman, N. D. 2017. Dissent: Sentence representation learning from explicit discourse relations. *arXiv preprint arXiv*:1710.04334.

Peters, M. E.; Neumann, M.; Iyyer, M.; Gardner, M.; Clark, C.; Lee, K.; and Zettlemoyer, L. 2018. Deep contextualized word representations. In *NAACL*, 2227–2237.

Radford, A.; Narasimhan, K.; Salimans, T.; and Sutskever, I. 2018. Improving language understanding by generative pre-training.

Rajpurkar, P.; Zhang, J.; Lopyrev, K.; and Liang, P. 2016. SQuAD: 100,000+ questions for machine comprehension of text. In *EMNLP*.

Socher, R.; Perelygin, A.; Wu, J. Y.; Chuang, J.; Manning, C. D.; Ng, A. Y.; Potts, C.; et al. 2013. Recursive deep models for semantic compositionality over a sentiment treebank. In *Proceedings of the conference on empirical methods in natural language processing (EMNLP)*, volume 1631, 1642.

Subramanian, S.; Trischler, A.; Bengio, Y.; and Pal, C. J. 2018. Learning general purpose distributed sentence representations via large scale multi-task learning. In *ICLR*.

Vaswani, A.; Shazeer, N.; Parmar, N.; Uszkoreit, J.; Jones, L.; Gomez, A. N.; Kaiser, Ł.; and Polosukhin, I. 2017. Attention is all you need. In *Advances in Neural Information Processing Systems*, 5998–6008.

Wang, A.; Singh, A.; Michael, J.; Hill, F.; Levy, O.; and Bowman, S. R. 2018. Glue: A multi-task benchmark and analysis platform for natural language understanding. *arXiv preprint arXiv:1804.07461*.

Williams, A.; Nangia, N.; and Bowman, S. R. 2017. A broadcoverage challenge corpus for sentence understanding through inference. *arXiv preprint arXiv*:1704.05426

Yang, Z.; Dai, Z.; Yang, Y.; Carbonell, J. G.; Salakhutdinov, R.; & Le, Q. V. 2019. XLNet: Generalized autoregressive pretraining for language understanding.